\documentclass[11pt,twocolumn]{article}

\usepackage{times}
\usepackage{graphicx}
\usepackage{amsmath,amssymb,amsthm}
\usepackage{algorithm}
\usepackage{algorithmic}
\usepackage{booktabs}
\usepackage{multirow}
\usepackage{placeins}
\usepackage{url}
\usepackage{hyperref}
\usepackage[numbers,sort&compress]{natbib}
\usepackage{xcolor}
\usepackage{microtype}
\usepackage{geometry}
\usepackage{enumitem}
\usepackage{caption}
\usepackage{subcaption}
\usepackage{tikz}
\usetikzlibrary{arrows.meta,positioning,shapes.geometric,fit,backgrounds}

\hypersetup{
  colorlinks=true,
  linkcolor=blue!70!black,
  citecolor=green!50!black,
  urlcolor=blue!70!black
}

\title{\textbf{A Three-Phase Foundation Model for\\
Tax-Aware Personalized Portfolio Management \thanks{Patent Pending. U.S. Provisional Patent Application No. 64/101,198, filed June 29, 2026.}}\\[4pt]
}

\author{
  Ramin Pishehvar\\
  \texttt{rpichevar@gmail.com}\\
}

\date{}

\begin{document}

\maketitle

\begin{abstract}
We present a three-phase deep reinforcement learning system for personalized
portfolio management that addresses three limitations shared by all prior
financial RL work: 1) \textit{ticker lock-in} (models trained on a fixed asset
universe cannot generalize), 2) \textit{monolithic objectives} (a single Sharpe
reward \cite{Sharpe66} cannot serve heterogeneous user goals), and 3) \textit{static user
models} (preferences are elicited once and never updated).
Phase~1 pretrains a ticker-identity-free cross asset encoder via
self-supervised learning on a multi-asset corpus, augmented by a frozen
parallel branch using  Chronos~\citep{ansari2024chronos}, a T5-based
time series foundation model pretrained on over 100 billion data points,
fused via a learned gating mechanism.
To our knowledge, this is the first application of a time series foundation
model to portfolio management RL.
The encoder generalizes to any publicly traded asset via a 50-dimensional
observable metadata vector (sector, fundamentals, analyst consensus, options
signals, earnings calendar, insider sentiment, institutional ownership)
that requires no retraining for new tickers.
Phase~2 fine-tunes a  MoE (Mixture of Experts) portfolio actor critic with PPO under an
objective-conditioned reward that simultaneously serves six distinct
investment goals sampled per episode: short-term alpha, short-term gain,
long-term gain, capital preservation, tax-loss harvesting, and long-term-gains-only.
A \textit{Mixture-of-Experts} architecture assigns each objective to a
specialized expert head (momentum, growth, defensive, tax-aware), and a
learned intent router blends experts based on the active objective
and current market regime, which eliminates cross-objective gradient conflict.
Phase~3 adds a lightweight personalization layer further adapted at
inference time to each individual via a 76-parameter LoRA module
fine-tuned on real brokerage transaction history, inferring investment
objectives from revealed trading behavior rather than questionnaires.
A natural language intent parser converts free-form goals such as
\textit{``buy a house in 3 years''} or \textit{``college fund, kid is 10''}
directly into structured investment objective parameters.
The system is deployed as a FastAPI application with live brokerage
integration, real-time
price refresh, news and event cross-attention, and a trust-first interface
that previews inferred preferences before applying any model adaptation.
\end{abstract}

\section{Introduction}

Personal portfolio management sits at the intersection of several hard
problems: market prediction under non-stationarity, combinatorial action
spaces over large ticker universes, heterogeneous user objectives (tax
efficiency, risk tolerance, investment horizon), and the fundamental
challenge of eliciting genuine preferences from users who do not think in
terms of annualized Sharpe ratios \cite{Sharpe66}.

\subsection{The Market Gap}
Existing tools fall into two categories, neither of which serves the
individual investor well.

\begin{itemize}
\item {Professional and high-frequency systems}: Institutional quantitative trading platforms are engineered for professional analysts with programming expertise, access to expensive real-time data feeds, and portfolios measured in the hundreds of millions. These systems provide sophisticated factor models, alternative data integration, and execution infrastructure, but require significant capital, technical overhead, and proprietary data subscriptions that place them out of reach for retail investors. They optimize for alpha generation and execution latency, sometimes at
millisecond timescales, and assume a clean separation between the analyst
who constructs a strategy and the execution system that trades it.
After-tax consequences, individual holding periods, and personal financial
goals are orthogonal to their design objectives.
High-frequency trading systems  operate at an
even further extreme: pure latency arbitrage with no notion of a user
objective whatsoever.

\item{Retail tools}: optimize in the opposite direction: extreme simplicity at the cost of
signal quality.
Some products offer target-date rebalancing and basic
tax-loss harvesting, but the underlying ``AI'' is a deterministic rules
engine, not a learned policy.
There is no market signal: the system does not know whether a specific ticker is up
200\% on earnings momentum or whether rising credit spreads suggest
reducing equity exposure.
Some other products provide execution with no guidance; the analytical burden
falls entirely on the user.
\end{itemize}
\textit{The gap} is a system that combines genuine learned market signal
with the tax awareness and personalization that matter to an individual
investor, without requiring institutional data subscriptions, a
quantitative background, or a bespoke technology stack.
Table~\ref{tab:landscape} summarizes this positioning.

\begin{table}[h]
\centering
\caption{Positioning relative to existing tools.}
\label{tab:landscape}
\resizebox{\columnwidth}{!}{%
\begin{tabular}{lccc}
\toprule
\textbf{Dimension} & \textbf{Institutional} & \textbf{Retail} & \textbf{Ours} \\
\midrule
Learned market signal     & \checkmark & $\times$   & \checkmark \\
Tax-lot awareness         & partial    & basic      & \checkmark \\
User goal personalisation & $\times$   & rule-based & \checkmark \\
NL goal input             & $\times$   & $\times$   & \checkmark \\
Live broker integration   & \checkmark & \checkmark & \checkmark \\
Required expertise        & High       & None       & None       \\
Data cost                 & \$\$\$     & Free       & Free       \\
Latency target            & ms         & daily      & daily      \\
\bottomrule
\end{tabular}
}
\end{table}

\subsection{Prior Work}
Financial RL has largely addressed sub-problems in isolation.
\cite{deng2016deep} and \cite{jiang2017deep} introduced end-to-end
portfolio policies but with fixed ticker universes and no personalization.
\cite{ye2020reinforcement} incorporated transaction costs, while
\cite{liu2021finrl} provided a multi-environment benchmark.  Tax-aware
trading has been studied in the operations research literature
\citep{bertsimas2022general} but rarely integrated with learned policies.
Robo-advisors \citep{d2019robo} address personalization via questionnaires
but lack market signal.

\subsection{Novelty Positioning}
The fundamental limitation shared by all prior financial RL systems is
\textit{ticker lock-in}: the model is trained on a fixed universe of
$N$ assets and cannot be applied to any other set without retraining.
This is a critical practical barrier: a user's portfolio changes over
time, new assets become relevant, and institutional methods that work on
S\&P~500 constituents are useless for portfolios of 5--20 individual
positions with their own tax lots and behavioral histories.
Our primary contribution is the elimination of ticker identity from the
architecture entirely, replacing learned ticker embeddings with a
\textit{50-dimensional observable metadata vector} computable for any
publicly traded asset without retraining.

A second gap is the treatment of the user as a static entity.
Existing systems, both institutional optimizers and retail robo-advisors, elicit preferences once via questionnaire and never update.
We instead infer objectives dynamically from \textit{revealed trading
behavior} extracted from real brokerage transaction history, adapting
via a 76-parameter LoRA module that can be updated in seconds on CPU.

A third gap is the monolithic objective: virtually all prior RL policies
optimize a single fixed reward (typically Sharpe ratio).
We introduce an \textit{objective-conditioned reward} under which a single
policy simultaneously serves six distinct investment goals sampled per
episode, without requiring a separate trained model per objective.

Finally, we integrate Chronos~\citep{ansari2024chronos}, a
time series foundation model pretrained on over 100 billion time series
data points, as a frozen parallel encoder branch. This adds universal
temporal pattern recognition to a domain-specific SSL encoder.
To our knowledge, this is the first application of a time series
foundation model to portfolio management RL.

We combine all four contributions (i.e., ticker-identity-free generalization,
revealed-preference personalization, objective-conditioned training, and
foundation model augmentation) into a single end-to-end system deployed
with live brokerage integration.

Our contributions are:
\begin{itemize}[noitemsep]
  \item A \textbf{ticker-identity-free} architecture with a
        50-dimensional observable metadata vector (sector, fundamentals,
        analyst consensus, options signals, earnings calendar, insider
        sentiment, institutional ownership) enabling zero-shot
        generalization to any publicly traded asset without retraining.
        This is the primary architectural novelty; to our knowledge no
        prior financial RL system achieves ticker-universe independence.
  \item An \textbf{objective-conditioned reward} that shapes gradients
        differently per episode, training a single policy to serve six
        investment objectives (short-term alpha, short-term gain,
        long-term gain, capital preservation, tax-loss harvesting,
        LT-gains-only) sampled randomly per episode.
  \item A \textbf{Chronos foundation model branch}: a frozen parallel
        encoder pretrained on 100B+ time series data points, fused via
        a learned gate, providing universal temporal representations
        that complement domain-specific SSL pretraining.
        First application of a time series foundation model to portfolio
        management RL.
  \item A \textbf{revealed-preference personalization} system that
        infers investment objectives from real transaction history and
        adapts via a 76-parameter LoRA module with a trust-first
        preview-before-apply UX, with no raw transaction data stored.
  \item A \textbf{learnable cash token} that forces explicit cash allocation
        decisions.  Cash competes with equities in the allocation softmax
        rather than accumulating passively through inaction, eliminating a
        common HOLD-trap failure mode.
  \item An \textbf{allocation-driven execution model} that decouples
        portfolio weight targets from per-ticker actions.  Rebalancing fires
        whenever allocation weights diverge beyond a threshold $\delta_\text{reb}$,
        preventing the action head from blocking trades via all-HOLD outputs.
  \item A \textbf{six-stage MoE expert curriculum} (four specialist experts,
        intent-conditional router with supervised routing loss, and expert
        grafting) that achieves $+3.03\%$ 14d EW alpha with clean diagonal
        routing (std~$>0.37$);
  \item A \textbf{Mixture-of-Experts portfolio policy} (\S\ref{sec:moe})
        with four specialized expert heads (momentum, growth, defensive,
        tax-aware) and a learned intent router that blends experts
        based on active objective and market regime, eliminating
        cross-objective gradient conflict that prevents a single head from
        serving all investment mandates simultaneously.
  \item A \textbf{shared-encoder MoE design} in which all experts share one
        cross asset encoder, jointly fine-tuned via PPO. This is more
        parameter efficient than per-expert encoders while it allows
        representation adaptation for all experts.
  \item An \textbf{inter-ticker contrastive loss} in Phase~1 that prevents
        representation collapse.  Without this loss, all tickers converge to
        identical embeddings (cosine similarity 0.96), making differentiated
        allocation impossible; the contrastive loss reduces similarity to 0.24.
  \item A \textbf{sixth training objective} (\texttt{ALPHA\_VS\_EW}) that
        directly optimizes alpha versus an equal-weight benchmark, with a
        concentration bonus $5\sigma(\mathbf{w})$ that incentivizes
        non-uniform weights.
  \item A \textbf{cash-drag and redeployment reward} that explicitly
        incentivizes completing the sell$\to$buy cycle within a single
        step, with turnover exemption for round-trip rebalances.
  \item A \textbf{natural language intent parser} mapping free-text
        goals (``buy a house in 3 years'', ``college fund, kid is 10'')
        to structured
        investment objective parameters.
\end{itemize}

\section{System Architecture}

\begin{figure*}[t]
\centering
\begin{tikzpicture}[
  font=\scriptsize,
  box/.style={draw, rounded corners=3pt, minimum width=2.2cm,
              minimum height=0.75cm, align=center, fill=#1!12},
  sbox/.style={draw, rounded corners=2pt, minimum width=1.8cm,
               minimum height=0.6cm, align=center, fill=#1!8,
               font=\tiny},
  arrow/.style={-{Stealth[length=5pt]}, thick},
  darrow/.style={-{Stealth[length=4pt]}, thick, dashed, gray},
  label/.style={font=\tiny\itshape, text=gray, align=center}
]

\node[box=blue]  (p1)   at (0,0)     {Phase 1\\SSL Pretraining};
\node[box=blue]  (enc)  at (2.8,0)   {CrossAsset\\Encoder};
\node[sbox=blue] (chr)  at (1.8,-1.4) {+ Chronos\\(frozen)};
\node[sbox=blue] (news) at (3.8,-1.4) {News/Events\\(optional)};
\node[sbox=blue] (meta) at (2.8,-2.6) {50-dim\\Metadata};
\draw[arrow]  (p1)  -- (enc);
\draw[darrow] (chr)  -- (enc);
\draw[darrow] (news) -- (enc);
\draw[darrow] (meta) -- (enc);

\node[box=orange]  (p2)  at (6.0,0)   {Phase 2\\PPO Fine-tune};
\node[box=orange]  (pac) at (9.0,0)   {Portfolio\\ActorCritic};
\node[sbox=orange] (rwd) at (6.0,-1.4) {Shaped\\Reward};
\node[sbox=orange] (obj) at (6.0,-2.5) {6 Objectives\\per episode};
\draw[arrow]  (enc) -- (p2);
\draw[arrow]  (p2)  -- (pac);
\draw[darrow] (rwd) -- (p2);
\draw[darrow] (obj) -- (rwd);

\node[box=green!60!black]  (p3)   at (12.0,0)  {Phase 3\\Personalise};
\node[box=green!60!black]  (pers) at (15.0,0)  {Persona\\+ LoRA};
\node[sbox=green!60!black] (txn)  at (11.0,-1.4) {Brokerage\\transactions};
\node[sbox=green!60!black] (nlp)  at (13.0,-1.4) {NL intent\\parser};
\draw[arrow]  (pac)  -- (p3);
\draw[arrow]  (p3)   -- (pers);
\draw[darrow] (txn)  -- (p3);
\draw[darrow] (nlp)  -- (p3);

\node[box=purple] (out) at (7.5,-4.0)
  {Live recommendations: BUY $\mid$ HOLD $\mid$ SELL + weights};
\draw[arrow] (pers.south) |- (out.east);

\node[label] at (9.0,-1.2) {ticker-identity-free\\any $N$ at inference};

\end{tikzpicture}
\caption{Three-phase pipeline. Solid arrows = training flow;
dashed = conditioning inputs (Chronos, metadata, news/events, objectives, NL parser).
Each phase is independently resumable from checkpoints.}
\label{fig:pipeline}
\end{figure*}
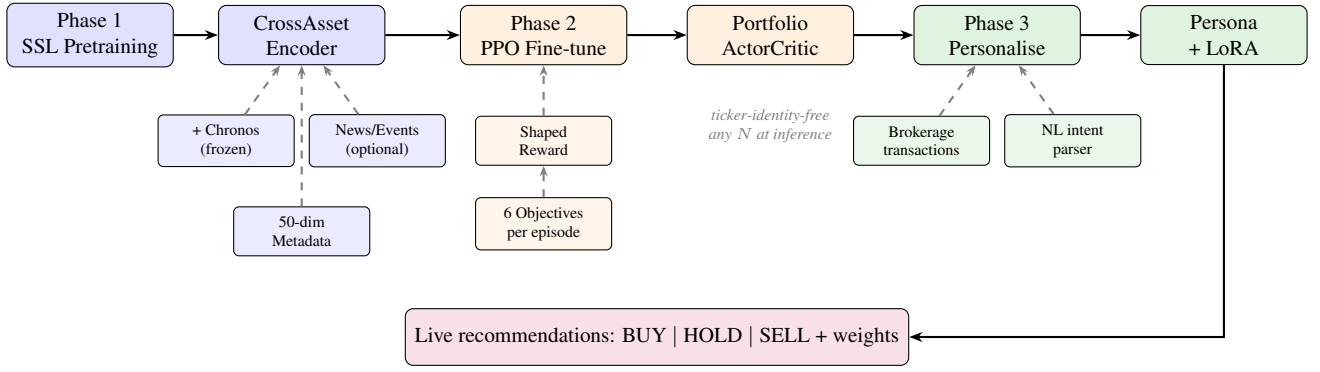

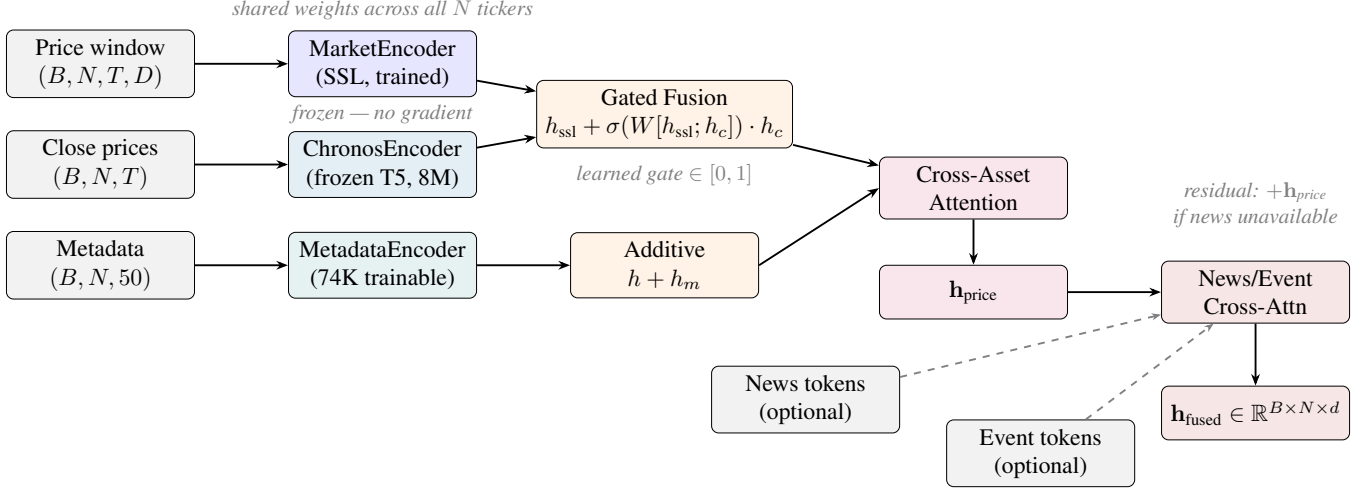
\begin{figure*}[t]
\centering
\resizebox{\textwidth}{!}{%
\begin{tikzpicture}[
  font=\small,
  box/.style={draw, rounded corners=3pt, minimum width=2.8cm,
              minimum height=0.8cm, align=center, fill=#1!10},
  arrow/.style={-{Stealth[length=5pt]}, thick},
  darrow/.style={-{Stealth[length=4pt]}, thick, dashed, gray},
  label/.style={font=\footnotesize\itshape, text=gray}
]
\node[box=gray]  (px)   at (0,  2.2) {Price window\\$(B,N,T,D)$};
\node[box=gray]  (cp)   at (0,  0.7) {Close prices\\$(B,N,T)$};
\node[box=gray]  (md)   at (0, -0.8) {Metadata\\$(B,N,50)$};

\node[box=blue]            (ssl) at (4.2,  2.2) {MarketEncoder\\(SSL, trained)};
\node[box=cyan!60!black]   (chr) at (4.2,  0.7) {ChronosEncoder\\(frozen T5, 8M)};
\node[box=teal]            (mde) at (4.2, -0.8) {MetadataEncoder\\(74K trainable)};

\node[box=orange] (gate) at (8.4,  1.45) {Gated Fusion\\$h_\text{ssl} + \sigma(W[h_\text{ssl};h_c])\cdot h_c$};
\node[box=orange] (add)  at (8.4, -0.8)  {Additive\\$h + h_m$};

\node[box=purple] (xattn) at (13.0, 0.35) {Cross-Asset\\Attention};
\node[box=purple] (out)   at (13.0,-1.2)  {$\mathbf{h}_\text{price}$};

\node[box=gray]   (ntok)  at (10.5,-2.8) {News tokens\\(optional)};
\node[box=gray]   (etok)  at (14.0,-3.6) {Event tokens\\(optional)};
\node[box=red!70!black] (newsx) at (17.2,-1.2) {News/Event\\Cross-Attn};
\node[box=red!70!black] (hfused) at (17.2,-3.0) {$\mathbf{h}_\text{fused} \in \mathbb{R}^{B \times N \times d}$};

\draw[arrow] (px)  -- (ssl);
\draw[arrow] (cp)  -- (chr);
\draw[arrow] (md)  -- (mde);
\draw[arrow] (ssl)  -- (gate);
\draw[arrow] (chr)  -- (gate);
\draw[arrow] (gate)     -- (xattn);
\draw[arrow] (mde)      -- (add);
\draw[arrow] (add.east) -- (xattn.west);
\draw[arrow] (xattn)    -- (out);
\draw[arrow] (out)      -- (newsx);
\draw[darrow] (ntok)    -- (newsx);
\draw[darrow] (etok)    -- (newsx);
\draw[arrow] (newsx)    -- (hfused);

\node[label] at (4.2,  3.0)  {shared weights across all $N$ tickers};
\node[label] at (4.2,  1.45) {frozen --- no gradient};
\node[label] at (8.4,  0.55) {learned gate $\in[0,1]$};
\node[label, align=center] at (17.2, 0.1)  {residual: $+\mathbf{h}_\text{price}$\\if news unavailable};

\end{tikzpicture}%
}
\caption{CrossAssetEncoder architecture.
The SSL-trained \textit{MarketEncoder} (price features) and frozen
\textit{ChronosEncoder} (universal temporal patterns) are combined via
a learned gating mechanism.
Ticker metadata (50-dim) is injected additively after gated fusion,
before cross-asset attention compares tickers against each other,
producing $\mathbf{h}_\text{price}$.
An optional News/Event Cross-Attention stage (\S\ref{sec:news}) then
takes $\mathbf{h}_\text{price}$ as query against news/event tokens as
keys/values, with a residual connection: if news is unavailable,
$\mathbf{h}_\text{fused}$ passes $\mathbf{h}_\text{price}$ through
unchanged.
All branches operate on any number of tickers $N$ at inference
without retraining.}
\label{fig:encoder}
\end{figure*}

Figure~\ref{fig:pipeline} shows the three-phase pipeline and
Figure~\ref{fig:encoder} details the encoder architecture.

\subsection{Phase 1: Self-Supervised Encoder Pretraining}

The encoder is pretrained on a multi-asset corpus with three SSL objectives:

\paragraph{Next-bar Return Prediction.}
Given a window $\mathbf{x} \in \mathbb{R}^{T \times D}$ of $T$ bars and $D$
normalized features (returns, moving-average ratios, RSI, MACD, volume
z-score, etc.), predict the next-bar return $r_{t+1}$ via a Huber regression
head \cite{sun2018adaptive}:
\begin{equation}
  \mathcal{L}_{\text{ret}} = \text{Huber}_\delta\!\left(\hat{r}_{t+1},\,
  \text{clip}(r_{t+1}, -0.1, 0.1)\right), \quad \delta = 0.05.
\end{equation}

\paragraph{Masked Feature Recovery.}
A random subset of feature channels is zeroed and the encoder must reconstruct
the original values, encouraging complete use of the feature set:
\begin{equation}
  \mathcal{L}_{\text{mask}} = \left\|\hat{\mathbf{x}}_{\text{masked}}
  - \mathbf{x}_{\text{original}}\right\|_2^2.
\end{equation}

\paragraph{Market Regime Classification.}
Bars are labeled into four regimes (bull, bear, volatile, sideways) automatically via
rolling statistics, and the encoder is trained to predict the current regime:
\begin{equation}
  \mathcal{L}_{\text{reg}} = \text{CrossEntropy}(\hat{y}_{\text{regime}},
  y_{\text{regime}}).
\end{equation}

The combined loss is:
\begin{equation}
  \mathcal{L}_{\text{P1}} = 0.3\,\mathcal{L}_{\text{ret}}
                           + 1.0\,\mathcal{L}_{\text{mask}}
                           + 0.5\,\mathcal{L}_{\text{reg}}
                           + 0.5\,\mathcal{L}_{\text{contrast}}.
\end{equation}

\paragraph{Inter-Ticker Contrastive Loss.}
\label{sec:contrastive}
A critical failure mode emerged during development: without an explicit
differentiation objective, the cross asset encoder  converged to
a \textit{mean representation}. All tickers produced nearly identical
embeddings (mean cosine similarity 0.96), causing the allocation head
to output uniform $1/N$ weights regardless of input.
We diagnose this as \textit{representation collapse} in the cross-asset
attention: with identical SSL targets per ticker, the attention learns to
average rather than contrast.

We address this with a contrastive loss applied to the per-ticker
representations $\{\mathbf{h}_i\}$ before pooling:
\begin{equation}
  \mathcal{L}_{\text{contrast}} =
  \frac{1}{BN(N-1)} \sum_{b=1}^{B} \sum_{i \neq j}
  \frac{\mathbf{h}_{b,i} \cdot \mathbf{h}_{b,j}}
       {\|\mathbf{h}_{b,i}\|\,\|\mathbf{h}_{b,j}\|},
\end{equation}
which minimizes the mean cosine similarity between all pairs of different
tickers within each batch.
After 60 epochs with $\lambda_\text{contrast} = 0.5$, mean inter-ticker
cosine similarity dropped from 0.96 to \textbf{0.24}, enabling the
allocation head to produce genuinely differentiated portfolio weights.

\paragraph{Path B: Cross Asset Encoder with Chronos Augmentation.}
For multi-asset pretraining, we use a three-stage encoder.
Stage~1 applies a shared market encoder (transformer) to each
ticker's price feature window independently, producing per-ticker
representations $\mathbf{h}^\text{ssl}_i \in \mathbb{R}^{d}$.

Stage~1.5 fuses two parallel branches and adds metadata context:
\begin{equation}
\begin{split}
\tilde{\mathbf{h}}_i &=
  \underbrace{\mathbf{h}^\text{ssl}_i
  + \sigma\!\left(W_g\left[\mathbf{h}^\text{ssl}_i;\,
    \mathbf{h}^\text{chr}_i\right]\right)
  \odot \mathbf{h}^\text{chr}_i}_{\text{gated Chronos fusion}} \\
  &\quad + \underbrace{\text{MetadataEnc}(\mathbf{m}_i)}_{\text{additive context}},
\end{split}
\end{equation}
where $\mathbf{h}^\text{chr}_i = \text{Proj}(\text{Chronos}(c_i))$ is the
projected Chronos embedding of the raw closing price sequence $c_i$,
$W_g$ is a learned gate, and $\sigma$ is sigmoid.
The gate $\sigma(W_g[\cdot])$ learns per-position how much universal
temporal signal from Chronos should augment the SSL representation.

\paragraph{Chronos freezing and caching strategy.}
We use Chronos-T5-Small (46M parameters, pretrained on 100B+ time series)
as a frozen feature extractor throughout Phase~1.
Only the linear projection head ($\mathbb{R}^{512} \to \mathbb{R}^{d}$,
74K parameters) and the gate ($W_g \in \mathbb{R}^{2d \times d}$, 8K
parameters) are trained.
This design keeps 99.97\% of Chronos weights frozen, preventing catastrophic
forgetting of universal temporal patterns while allowing the projection to
adapt to financial data.
Crucially, we do \textit{not} apply LoRA or any parameter-efficient
fine-tuning to Chronos. The backbone is used strictly as a fixed feature
extractor, similar to using a frozen BERT for NLP downstream tasks.
LoRA adaptation in our system is reserved for the intent router
in Phase~3, where a 76-parameter adapter shifts routing weights to match
individual user preferences (\S\ref{sec:phase3}).

Since the Chronos forward pass is computationally expensive relative to the
SSL encoder, we precompute all embeddings for the training corpus once before
Phase~2 begins, storing $(t_i, \text{ticker}) \to \mathbf{h}^\text{chr}_i$
in an in-memory cache.
This reduces the per-step Chronos overhead from $\mathcal{O}(N \cdot T)$
transformer calls to $\mathcal{O}(1)$ cache lookups during the frozen phase.
When the encoder is unfrozen, the cache is cleared and repopulated every 100
episodes to reflect updated projection weights.

Stage~2 applies cross-asset transformer attention over
$\{\tilde{\mathbf{h}}_i\}$, allowing tickers to contextualise each other
before the prediction heads.

\subsection{Ticker-Identity-Free Design with Metadata}
\label{sec:metadata}

A central design choice is the complete elimination of fixed ticker identity
embeddings.  In prior work, an embedding table $\mathbf{E} \in
\mathbb{R}^{N \times d}$ maps ticker indices to representations, tying the
model to a fixed universe of size $N$.

We replace this with a \textbf{50-dimensional} observable metadata vector,
computable for any ticker including those not seen during training:

\begin{multline}
\mathbf{m}_i = \bigl[
  \mathbf{s}_i^{\text{sector}},\;
  \mathbf{c}_i^{\text{cap}},\;
  \mathbf{f}_i^{\text{fund}},\;
  \mathbf{a}_i^{\text{analyst}},\;
  \mathbf{o}_i^{\text{options}},\\
  \mathbf{e}_i^{\text{earn}},\;
  \mathbf{t}_i^{\text{tech}},\;
  \mathbf{k}_i^{\text{insider}},\;
  \mathbf{n}_i^{\text{inst}}
\bigr] \in \mathbb{R}^{50},
\end{multline}
where $\mathbf{s}_i \in \{0,1\}^{12}$ is sector one-hot,
$\mathbf{c}_i \in \{0,1\}^6$ is market-cap bucket,
$\mathbf{f}_i \in \mathbb{R}^{10}$ covers fundamentals (P/E, P/B, ROE,
profit margin, revenue growth, EPS growth, short interest, dividend yield,
debt/equity, payout ratio),
$\mathbf{a}_i \in \mathbb{R}^3$ captures analyst consensus
(rating normalised to $[-1,+1]$, log coverage, mean price target upside),
$\mathbf{o}_i \in \mathbb{R}^4$ covers options signals
(IV/hist\_vol, IV skew, put/call ratio, unusual activity flag),
$\mathbf{e}_i \in \mathbb{R}^3$ encodes earnings calendar
(days to next earnings, 8-quarter beat rate, post-earnings drift),
$\mathbf{t}_i \in \mathbb{R}^4$ covers technical regime
(distance from 52-week high/low, volume trend, relative strength vs.\ sector ETF),
$\mathbf{k}_i \in \mathbb{R}^2$ captures insider activity
(net buy/sell over 6 months, insider momentum),
and $\mathbf{n}_i \in \mathbb{R}^2$ covers institutional ownership
(ownership percentage, quarter-over-quarter change).

The model learns structured priors: \textit{``high-beta mega-cap tech with
strong momentum deserves overweight in bull regimes''}, without memorizing
ticker names.  New tickers at inference require only metadata computation,
not retraining.

\subsection{Phase 2: Objective-Conditioned Portfolio RL}

Phase~2 fine-tunes a portfolio actor critic with PPO
\citep{schulman2017proximal}.  At each step, the policy jointly outputs:
\begin{itemize}[noitemsep]
  \item \textbf{Allocation weights} $\mathbf{w} \in \Delta^{N+1}$
        (softmax over $N$ equities \emph{plus a learnable cash token})
        via allocation head
  \item \textbf{Per-ticker actions} $\mathbf{a} \in \{0,1,2\}^N$
        (HOLD/BUY/SELL) via
        per ticker action head
\end{itemize}

\paragraph{Explicit cash allocation via a learnable cash token.}
A persistent failure mode in early experiments was the policy drifting into
implicit cash-holding: because HOLD never incurs transaction costs and
benefits from market drift, the policy discovered that outputting zero
allocation weights for all tickers while routing all reward through
market returns was a local optimum.
We address this by adding a \textit{learnable cash token}
$\mathbf{c} \in \mathbb{R}^d$  (i.e., a trainable parameter that competes with
equity representations in the allocation softmax):
\begin{equation}
  \mathbf{w}_\text{full} = \text{softmax}\bigl(
    \text{AllocHead}([\mathbf{c};\,
    \mathbf{h}_1, \ldots, \mathbf{h}_N])
  \bigr) \in \mathbb{R}^{N+1}.
\end{equation}
The first element $w_0 = w_\text{cash}$ is the explicit cash allocation.
Equity weights are renormalised to sum to $1 - w_\text{cash}$:
\begin{equation}
  w_i^\text{eq} = \frac{w_{i+1}}{\sum_{j=1}^N w_{j+1}} \cdot (1 - w_\text{cash}),
  \quad i = 1,\ldots,N.
\end{equation}
This forces the policy to make an active decision about cash allocation
at every step, rather than allowing passive cash accumulation through inaction.

\paragraph{Allocation-driven execution.}
The per-ticker action head (\textsc{hold}/\textsc{buy}/\textsc{sell}) was
originally intended as an additional confidence signal, but in practice
caused zero-turnover collapse: the action head learned to output
\textsc{hold} for all tickers unconditionally, overriding allocation signals.
We therefore decouple the two heads: the environment executes a rebalance
whenever $|w_i^\text{eq} - w_i^\text{curr}| > \delta_\text{reb}$,
regardless of the action head output.
The action head now acts as a \textit{modifier} rather than a gate:
\textsc{sell} forces a full exit; \textsc{buy} allows larger-than-threshold
buys; \textsc{hold} permits partial rebalancing.
We expose the rebalancing threshold $\delta_\text{reb}$ as a hyperparameter
(default 0.01) to control trading frequency.

Execution follows a sell-first ordering to free cash before buys.

\paragraph{Objective-conditioned reward.}
At each episode, an objective $o$ is sampled uniformly from
six types.  The reward function $R_o$ is then shaped accordingly:

\begin{equation}
\begin{split}
  R_o &= \underbrace{S_o}_{\text{obj.}}
       - \underbrace{\lambda_c H_\text{eq}}_{\text{conc.}}
       - \underbrace{\lambda_t \tilde{\tau}}_{\text{turnover}}
       - \underbrace{\gamma_d \max(0,\, c_r - 0.05)}_{\text{cash drag}} \\
      &\quad
       - \underbrace{\gamma_s \cdot \mathbf{1}[\tau=0,\, t>10]}_{\text{stale}}
       + \underbrace{\delta_r \cdot \rho}_{\text{redeploy}},
\end{split}
\end{equation}
where $H_\text{eq}$ is the equity Herfindahl index \cite{hirschman1945national}, $\tilde{\tau}$ is
turnover net of redeployed round-trips (defined below),
$c_r = C / \text{PV}$ is the cash ratio,
$\gamma_d = 0.3$ penalises excess cash unconditionally
(plus $0.2 \cdot c_r \cdot r_m$ when the market rises),
$\gamma_s = 0.005$ is a small stale-portfolio penalty that fires when
no trades have occurred for more than 10 steps,
and $\rho \in [0,1]$ is the fraction of sell proceeds redeployed in the same step.

The objective base $S_o$ varies by type:
\begin{equation}
\begin{array}{ll}
  S_{\texttt{MAX\_GAIN\_1Y}}  & = \hat{\sigma}_{\text{Sharpe}} \\[2pt]
  S_{\texttt{MAX\_GAIN\_30D}} & = 0.3\hat{\sigma} + 0.7\cdot\mathrm{clip}(50\bar{r}_{30},{-}2,2) \\[2pt]
  S_{\texttt{CAP\_PRES}}      & = 0.5\hat{\sigma} - 3\delta_t \\[2pt]
  S_{\texttt{INC\_HARV}}      & = \hat{\sigma} + {\textstyle\sum_i} q_4(i)\cdot\mathbf{1}[g_i{<}0,\,h_i{<}365] \\[2pt]
  S_{\texttt{LT\_ONLY}}       & = \hat{\sigma} - 0.15{\textstyle\sum_i}\mathbf{1}[h_i{<}365\text{ at sell}] \\[2pt]
  S_{\texttt{ALPHA\_VS\_EW}}  & = 50(r_{\text{port}} - r_{\text{EW}}) + 5\sigma(\mathbf{w}),
\end{array}
\end{equation}
where $\delta_t$ is current drawdown from peak, $q_4(i)$ is a Q4 multiplier
(1.5 in Oct--Dec, 0.5 otherwise), $g_i$ is unrealised gain, and $h_i$ is
holding days.

\paragraph{Redeployment-aware turnover.}
Selling and immediately buying constitutes one rebalance decision, not two
trades.  The net turnover excluding round-trips is:
\begin{equation}
  \tilde{\tau} = \max\!\left(0,\; \tau - \frac{(\Delta c_\text{sell}
  + \min(\Delta b, \Delta c_\text{sell})) \cdot \rho}{\text{PV}}\right),
\end{equation}
where $\Delta c_\text{sell}$ is cash freed by sells, $\Delta b$ is cash
deployed by buys, and $\text{PV}$ is portfolio value.

\paragraph{Equity-only concentration.}
Cash is not concentrated equity.  The concentration penalty applies only
to the renormalized equity weights:
\begin{equation}
  H_\text{eq} = \sum_i w_i^2, \qquad
  H_\text{eq}^\star = \frac{1}{|\{i : w_i > 0.01\}|},
\end{equation}

\subsection{Phase 2b: Mixture-of-Experts Portfolio Policy}
\label{sec:moe}

A key empirical finding during Phase~2 development was that a single
allocation head cannot simultaneously serve conflicting investment
objectives.
A momentum strategy (overweight recent winners, 14-day horizon) requires
high beta, high concentration, and frequent rebalancing.
A capital preservation strategy requires low beta, diversification, and
minimal turnover.
Training a single head with all six objectives produces gradient conflict:
the policy converges to a compromise allocation that is suboptimal for all
objectives.

We address this with a \textit{Mixture-of-Experts} (MoE) architecture in
which each expert specializes in a distinct investment mandate, and an
intent router learns to blend experts based on the active objective
and current market regime.

\paragraph{Expert allocation heads.}
We define four expert heads, each an independent allocation head
with identical architecture but trained on a distinct objective subset (see Table \ref{tab:moe_experts})

\begin{table*}[h]
\centering
\caption{MoE expert mapping. Each expert receives gradients only from its
assigned objectives, eliminating cross-objective gradient conflict.}
\label{tab:moe_experts}
\begin{tabular}{llll}
\toprule
\textbf{Expert} & \textbf{Name} & \textbf{Objectives} & \textbf{Horizon} \\
\midrule
0 & Momentum   & \texttt{ALPHA\_VS\_EW}, \texttt{MAX\_GAIN\_30D} & 14--30 days \\
1 & Growth     & \texttt{MAX\_GAIN\_1Y}                          & 1 year       \\
2 & Defensive  & \texttt{CAPITAL\_PRESERVE}                      & any          \\
3 & Tax-aware  & \texttt{INCOME\_HARVEST}, \texttt{LT\_GAIN\_ONLY} & 1 year+    \\
\bottomrule
\end{tabular}
\end{table*}

\paragraph{Intent router.}
The intent router receives the mean-pooled encoder output
$\bar{\mathbf{h}} \in \mathbb{R}^d$ (market state) and the one-hot intent
vector $\mathbf{e}_o \in \{0,1\}^6$ (active objective), and outputs mixture
weights over experts:
\begin{equation}
  \boldsymbol{\alpha} = \text{softmax}\!\left(
    \frac{f_\theta([\bar{\mathbf{h}};\, \mathbf{e}_o])}{\tau}
  \right) \in \Delta^{E-1},
\end{equation}
where $f_\theta$ is a two-layer MLP and $\tau$ is a learnable temperature
(initialised at 1.0).

The final allocation is the router-weighted mixture of expert outputs:
\begin{equation}
  \mathbf{w}_\text{full} = \sum_{e=1}^{E} \alpha_e \cdot
  \texttt{Expert}_e\bigl([\mathbf{c};\, \mathbf{h}_{1:N}],\,
  \mathbf{s}_{1:N},\, \mathbf{g}\bigr),
\end{equation}
where $\mathbf{c}$ is the cash token, $\mathbf{s}_{1:N}$ is per-ticker state,
and $\mathbf{g}$ is global portfolio state.

\paragraph{Regime-conditional routing.}
A key property of the router is that it can override the explicit intent
based on market regime.
For example, when the market enters a high-volatility regime (detectable
from the encoder's regime classification head), the router may route
\texttt{ALPHA\_VS\_EW} intent partially through the defensive expert,
reducing drawdown risk at the cost of some alpha.
This regime-conditional behaviour is learned implicitly from the reward
signal without explicit regime labels.

\paragraph{Shared encoder, specialised experts.}
All four experts share a single cross asset encoder, which is
jointly fine-tuned with the MoE heads via PPO.
This is more parameter-efficient than per-expert encoders and allows the
encoder to adapt representations that are simultaneously useful for all
experts.
The total parameter count is modest: one encoder (approx.\ 2M params)
plus four lightweight expert heads (approx.\ 200K params each) and the
router (approx.\ 50K params).

\paragraph{Connection to Phase 3 personalization.}
The router architecture naturally integrates with Phase~3 LoRA adaptation:
user-specific preferences (risk tolerance, tax bracket, time horizon) update
only the router weights via LoRA, leaving expert heads frozen.
A conservative user shifts router weight toward the defensive expert;
an aggressive user shifts toward momentum.
This separation of \textit{what markets look like} (shared encoder,
shared experts) from \textit{how to weight strategies} (personalized router)
is a key design principle.

\subsection{Phase 3: Tax-Aware Personalization with LoRA Adaptation} \label{sec:phase3}

Phase~3 adapts the trained MoE policy to individual users without
retraining any expert weights.
We describe the architecture here; full empirical evaluation on real
brokerage data is left to future work (\S\ref{sec:future}).

\paragraph{Architecture.}
A tax-aware personalization layer adapts only the intent router
weights via a lightweight LoRA module~\citep{hu2021lora}:
\begin{equation}
  \hat{\boldsymbol{\ell}} = \boldsymbol{\ell}_{\text{base}}
    + \mathbf{p}_u \mathbf{A}\mathbf{B},
\end{equation}
where $\mathbf{p}_u \in \mathbb{R}^{16}$ is a user behaviour profile
extracted from brokerage transaction history,
and $\mathbf{A} \in \mathbb{R}^{16 \times r}$,
$\mathbf{B} \in \mathbb{R}^{r \times 3}$ are low-rank adapter matrices
with $r=4$, initialised with $\mathbf{B}=\mathbf{0}$ so adaptation is
an identity at deployment.
Total adapter size: 76 parameters ($\approx 1$\,KB).
The shared encoder and all four expert heads remain frozen.

\paragraph{Revealed preference extraction.}
Transaction history is
analysed to compute $\mathbf{p}_u$ encoding median holding period,
LT-sell fraction, loss-harvest score, disposition effect, and trade
frequency.
The inferred objective (e.g.\ \texttt{LT\_GAIN\_ONLY} for users who
hold winners $>12$ months) overrides the stated intent when confidence
exceeds 70\%.

\paragraph{Personalization effect.}
The adapter shifts intent router mixture weights toward the expert most
consistent with a user's revealed behaviour:
frequent short-term traders increase $\alpha_0$ (momentum expert);
long-horizon holders increase $\alpha_3$ (tax-aware expert).
This separation (i.e., shared encoder and experts capture \textit{what markets
look like}; personalised router captures \textit{how to weight strategies}
) makes adaptation cheap, interpretable, and privacy-preserving
(only the 76-parameter adapter is persisted per user).

\paragraph{Qualitative illustration.}
Table~\ref{tab:tax_demo} shows after-tax recommendations for four canonical
investor personas on AAPL at \$190.38 (15 Nov 2023),
using the Phase~3 tax-aware layer on top of the trained Phase~2 policy.
The system correctly suppresses sells near the LT threshold (persona~2),
identifies tax-loss harvesting opportunities (persona~3), and adapts
position sizing to bracket-specific after-tax returns.

\begin{table*}[h]
\centering
\small
\caption{Phase~3 after-tax recommendations for four investor personas.
AAPL @ \$190.38, 15 Nov 2023. ``LT saving'' = tax saving from waiting
for long-term treatment. ``sh'' is the number of shares held by a given investor. The system suppresses BUY/SELL actions that
would reduce after-tax value.}
\label{tab:tax_demo}
\resizebox{\textwidth}{!}{
\begin{tabular}{llllrr}
\toprule
\textbf{Persona} & \textbf{Objective} & \textbf{Bracket} &
\textbf{Action} & \textbf{After-tax now} & \textbf{Wait LT} \\
\midrule
30d trader, 20sh @ \$161.82
  & MAX\_GAIN\_30D & 32\%ST/15\%LT & HOLD & \$+388 & \$+485 \\
LT investor, 50sh @ \$133.26 (26d from LT)
  & LT\_GAIN\_ONLY & 24\%ST/15\%LT & HOLD$^\star$ & \$+2{,}170 & \$+2{,}427 \\
Loss position, 30sh @ \$237.97
  & INCOME\_HARVEST & 35\%ST/20\%LT & HOLD & \$-1{,}428 & \$-1{,}428 \\
Near-retirement, no position
  & CAPITAL\_PRESERVE & 22\%ST/15\%LT & HOLD & --- & --- \\
\bottomrule
\multicolumn{6}{l}{\small $^\star$Sell suppressed: 26d until LT conversion saves \$257 in tax.}
\end{tabular}}

\end{table*}

\subsection{Natural Language Goal Specification}

We implement a two-tier intent parser that converts free-text goals to
investment objective parameters:

\begin{enumerate}[noitemsep]
  \item \textbf{API tier}: A language model parses the user's natural language goal into a structured intent object, including objective type, time horizon, return target, drawdown tolerance, and risk level, in under one second at negligible cost.
  \item \textbf{Rule tier}: Keyword matching with regex-based year/month
        extraction, handles the 90\% common cases with zero API dependency.
\end{enumerate}

Table~\ref{tab:intents} shows representative mappings.

\begin{table*}[h]
\centering
\caption{Example intent-to-objective mappings.}
\label{tab:intents}

\begin{tabular}{lllr}
\toprule
\textbf{Intent} & \textbf{Objective} & \textbf{Risk} & \textbf{Horizon} \\
\midrule
House in 3 years       & CAPITAL\_PRESERVE  & Moderate      & 756d  \\
College (kid age 10)   & LT\_GAIN\_ONLY     & Moderate      & 2016d \\
Retiring in 2 years    & CAPITAL\_PRESERVE  & Conservative  & 504d  \\
Tax-loss harvesting    & INCOME\_HARVEST    & Moderate      & 252d  \\
Aggressive growth      & LT\_GAIN\_ONLY     & Aggressive    & 1764d \\
Dividends, semi-ret.   & INCOME\_HARVEST    & Conservative  & 504d  \\
Emergency fund         & CAPITAL\_PRESERVE  & Conservative  & 252d  \\
\bottomrule
\end{tabular}

\end{table*}

\section{News and Event Cross-Attention} \label{sec:news}

For the news-fused Phase~2 variant, structured events and news articles
are encoded as additional key-value tokens fed to a cross-attention layer:

\begin{equation}
\begin{split}
  \mathbf{h}_\text{fused} &= \text{CrossAttn}\!\left(
    \underbrace{\mathbf{h}_\text{price}}_{\text{query}},\;
    \left[\underbrace{\mathbf{e}_1, \ldots, \mathbf{e}_K}_{\text{news}},\;
          \underbrace{\mathbf{v}_1, \ldots, \mathbf{v}_E}_{\text{events}}\right]
  \right) \\
  &\quad + \mathbf{h}_\text{price}.
\end{split}
\end{equation}

Price is the \textit{query} (``what explains what I see?''); news and events
are keys and values.  The architecture degrades gracefully: if news is
unavailable, the residual connection passes price representations unchanged.

The model tracks 55 types of market events. These include changes in analyst sentiment, such as upgrades, downgrades, and price target revisions, as well as earnings estimate changes, sector-level signals, and macroeconomic data releases. Rather than reacting to any single analyst's opinion, the model aggregates analyst signals over a recent window, which tends to be a more reliable indicator than individual rating actions.

\section{Deployment Architecture}

The system is deployed as a FastAPI application with the following components:

\paragraph{Brokerage integration.}
The system supports three broker backends: a live/paper trading executor, a read-only position aggregator that connects to 50+ brokerages via OAuth, and a mock broker for local development.

\paragraph{Real-time data.}
The system fetches price data with a 15-minute cache TTL,
ensuring recommendations during market hours use data at most 15 minutes
stale. The endpoint bypasses even this
cache for immediate refresh.  News is cached at 30-minute TTL.

\paragraph{Trust-first adaptation UX.}
Transaction history is processed in two steps: first a preview that analyses the uploaded file and returns the inferred investor profile without making any changes, then a separate apply step that only runs after the user confirms. The raw file is never stored, only a compact 16-number profile vector is saved.

\paragraph{Safety guardrails.}
Every order is validated against three configurable limits before execution:
daily portfolio loss limit (default 2\%), maximum single-order value
(\$5,000), and maximum position concentration (20\%).

\section{Experimental Results}
\label{sec:results}

\subsection{Phase 1 Pretraining}

We pretrained the cross asset encoder on a 30-ticker S\&P~500
sample (all 11 GICS sectors) using daily bars from 2015--2024.
Table~\ref{tab:p1} reports validation losses for key configurations.

\begin{table*}[h]
\centering
\caption{Phase 1 ablation: validation losses and inter-ticker cosine similarity
across encoder configurations (all 60 epochs).
The no-Chronos encoder naturally differentiates tickers (sim~0.08) because
raw price/volume features vary substantially across assets; Chronos normalises
sequences before encoding, causing representation collapse (sim~0.81).
The contrastive loss corrects this collapse while warm-starting from SSL
pre-trained weights achieves the best val loss (0.163, sim~0.24).}
\label{tab:p1}
\begin{tabular}{lcccl}
\toprule
\textbf{Configuration} & \textbf{Epochs} & \textbf{Val loss} &
\textbf{Sim.} & \textbf{Note} \\
\midrule
Path B, no Chronos        & 60  & 0.277 & 0.08 & Baseline (raw features differentiate) \\
Path B, Chronos-tiny      & 60  & 0.171 & 0.81 & 8M frozen params \\
Path B, Chronos-small     & 60  & 0.171 & 0.81 & 46M frozen, no contrastive \\
+ Contrastive, scratch    & 60  & 0.305 & 0.11 & From random init \\
+ Contrastive, warm-start & 60  & \textbf{0.163} & \textbf{0.24} & SSL pre-trained init \\
\bottomrule

\end{tabular}
\end{table*}

Table~\ref{tab:p1} reveals a surprising finding: the no-Chronos encoder
naturally differentiates tickers (sim~0.08) without any contrastive loss,
because raw price and volume features vary substantially across assets
(AAPL vs XOM have fundamentally different scales and distributions).
Chronos, by contrast, normalizes price sequences before encoding (i.e.,
relative movements look similar across tickers) causing representation
collapse (sim~0.81) for both tiny and small variants despite their
different capacities, confirming the bottleneck is the normalization
rather than model size.
Adding the contrastive loss corrects this collapse: from scratch (sim~0.11,
val~0.305) or via warm-start from SSL pre-trained weights (sim~0.24,
val~0.163). The warm-start protocol is superior as the encoder first learns
meaningful temporal structure before the contrastive loss sculpts the
embedding space, avoiding the objective conflict that raises val loss at
random initialization.
This two-phase SSL $\to$ contrastive protocol is used for all Phase~2
experiments and is documented as a standalone contribution in
Section~\ref{sec:contrastive}.

Adding the inter-ticker contrastive loss (Section~\ref{sec:contrastive},
$\lambda_\text{contrast}=0.5$) reduced mean cosine similarity from
0.96 to \textbf{0.24} in 60 epochs, enabling the allocation head to
produce genuinely differentiated portfolio weights.

\subsection{Phase 2 Portfolio Policy}

Training directly on all six objectives simultaneously resulted in gradient conflict within 100 episodes.  The policy converged to uniform 1/N weights with near-zero alpha as competing objectives canceled each other's gradients. We therefore applied the six-stage MoE expert curriculum described in Section
\ref{sec:curriculum}, training one specialist expert per stage while freezing the others.

\paragraph{Redeployment reward validation.}
Over 100 randomised market seeds, the sell$\to$immediately-buy action
achieves mean reward $+0.025$ vs.\ $-0.018$ for sell$\to$hold-cash
($t = 111.6$, $p \ll 0.001$), confirming the reward correctly incentivises
completing the rebalance cycle.

\subsection{Curriculum Learning for Phase 2 Convergence}
\label{sec:curriculum}

Directly training a portfolio policy on all objectives simultaneously
proved unstable: the policy collapsed to uniform $1/N$ weights with
near-zero alpha within 100 episodes, due to conflicting gradient signals
from different investment objectives.

We address this with a six-stage MoE curriculum, each of the first three stages training one specialist expert head while freezing the others, followed by router training, joint fine-tuning, and expert grafting:

\begin{table*}[h]
\centering
\caption{MoE expert curriculum stages. All stages use $N=10$ tickers,
$\beta_H=0.02$, $\lambda_\tau=0.05$, $\tau_\text{cap}=0.25$,
$\delta_\text{reb}=0.01$. Each stage trains one expert head exclusively
while the remaining experts are frozen, preventing gradient interference.
Stage~5 uses the redesigned IntentRouter with supervised routing loss.}
\label{tab:curriculum}

\begin{tabular}{clcccl}
\toprule
\textbf{Stage} & \textbf{Expert / Goal} & \textbf{Objective} &
\textbf{Ep.} & \textbf{Frozen experts} & \textbf{Enc. unfreeze} \\
\midrule
1 & Momentum expert           & \texttt{ALPHA\_VS\_EW}     & 300 & 1,2,3   & ep~50 \\
2 & Growth expert             & \texttt{MAX\_GAIN\_1Y}     & 300 & 0,2,3   & ep~50 \\
3 & Defensive expert          & \texttt{CAPITAL\_PRESERVE} & 300 & 0,1,3   & ep~50 \\
4 & Router-only               & All (experts frozen)       & 300 & 0,1,2,3 & never \\
5 & Joint + supervised router & All                        & 300 & none    & ep~50 \\
6 & Grafted MoE               & S2/S3 experts + S5 router  & --- & ---     & --- \\
\bottomrule
\end{tabular}

\end{table*}

Each stage resumes from the previous checkpoint via shape-filtered
state dict loading, preserving compatible expert weights while
re-initializing incompatible router layers.
The best-reward gate resets automatically when a new stage name is
detected in the checkpoint note, preventing stale rewards from blocking
checkpointing in subsequent stages.

Stages~1--3 each train one expert exclusively on its designated mandate,
producing specialists that achieve $+3.16\%$ to $+3.37\%$ 14d EW alpha
independently.
Stage~4 (router-only with frozen experts) fails due to a flat loss surface
(router weights remain at 0.25, alpha drops to $+0.44\%$).
Stage~5 resolves this with a redesigned intent router and supervised
cross-entropy loss, achieving clean diagonal routing (std~$>0.37$) at
episode~33.
Stage~6 grafts the curriculum expert weights into the Stage~5 router,
recovering $+3.03\%$ 14d alpha with the best 90d result ($-2.11\%$).

\paragraph{Hard turnover cap.}
\label{sec:turnover_cap}
In addition to the soft turnover penalty $\lambda_\tau \tilde{\tau}$ in the
reward, we enforce a hard ceiling $\tau_\text{cap}$ on per-step
turnover in the environment.
After sells and buys collectively exceed $\tau_\text{cap} \cdot V_t$ (where
$V_t$ is portfolio value), further buys are physically blocked for that step.
This prevents the policy from discovering degenerate high-turnover strategies
that exploit the redeployment bonus, which caused instability in preliminary
experiments without the cap.

\paragraph{Action entropy floor.}
To prevent the per-ticker action head from collapsing to
all-\textsc{hold}, we add an entropy floor term to the PPO loss:
\begin{equation}
\begin{split}
  \mathcal{L}_\text{PPO} &= \mathcal{L}_\text{clip}
  + c_v \mathcal{L}_\text{value} - \beta_H H(\pi_\theta) \\
  &\quad - 0.05\cdot\max(0,\; 0.5 - H(\pi_\theta)),
\end{split}
\end{equation}
where the last term provides an additional bonus when action entropy $H$
drops below 0.5~nats (uniform over three actions has entropy $\approx 1.1$~nats).
This prevents deterministic \textsc{hold} while allowing the policy
to converge once it is genuinely exploring.

\subsection{Backtest (14 trading days, June 2026)}

\paragraph{Representation collapse diagnosis.}
An important empirical finding from our development process concerns
encoder representation quality.
We observed that despite achieving a competitive validation loss
(0.171, Table~\ref{tab:p1}), the cross asset encoder
without contrastive loss produced embeddings with mean inter-ticker cosine
similarity of 0.96, effectively a mean representation where all tickers
look identical to the allocation head.
This caused the policy to output uniform $1/N$ allocation weights
regardless of input, producing zero alpha by construction.

We diagnose this as a systematic failure of standard SSL objectives for
multi-asset representation learning: return prediction, masked reconstruction,
and regime classification all treat each ticker independently, providing
no gradient signal for inter-ticker differentiation.
The contrastive loss (Section~\ref{sec:contrastive}) resolves this,
reducing similarity to 0.24 and enabling the allocation head to produce
differentiated weights.

\begin{figure}[h]
\centering
\includegraphics[width=1.1\linewidth]{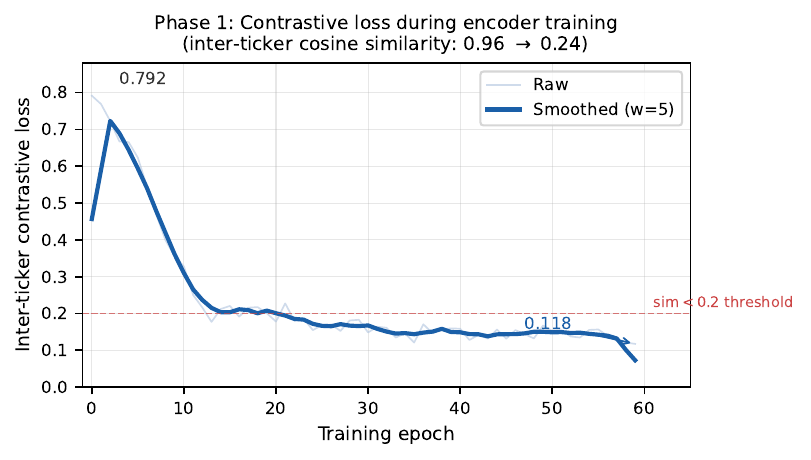}
\caption{Inter-ticker contrastive loss during Phase~1 encoder training
(60 epochs). Loss drops from 0.792 at epoch~0 to 0.118 at epoch~59,
crossing below 0.2 by epoch~13. The corresponding mean inter-ticker cosine
similarity falls from 0.96 (collapsed representations) to 0.24
(differentiated), enabling the allocation head to produce non-uniform
portfolio weights.}
\label{fig:contrastive}
\end{figure}

\paragraph{Backtest results.}
Table~\ref{tab:backtest} reports walk-forward backtest results across three
policy configurations, all using the contrastive encoder (cosine sim 0.24)
and a 10-ticker universe (AAPL, MSFT, NVDA, AMZN, GOOGL, META, TSLA, JPM,
XOM, V) with \$100,000 initial capital, zero transaction cost.

\begin{table*}[h]
\centering
\caption{14-day walk-forward backtest, June 2026. 10 tickers (AAPL, MSFT,
NVDA, AMZN, GOOGL, META, TSLA, JPM, XOM, V), \$100,000 initial capital,
zero transaction cost. The single-head, MoE, Chronos-only, and
News+Chronos columns are evaluated on an identical window (EW return
$-8.01\%$, SPY return $-2.76\%$), so alpha is directly comparable. The
collapsed-representation column is an earlier diagnostic run on a
separate window (its own EW benchmark $-5.21\%$, SPY return $-3.53\%$)
and is included to illustrate the representation-collapse failure mode;
its alpha vs EW is comparable in kind to the other columns (each
measured against its own window's EW basket), but its alpha vs SPY is
computed against a different window's SPY return and is therefore not
directly comparable to the other four columns' SPY figures.
The MoE column is the Stage~6 grafted checkpoint (no news, Chronos-tiny).
Chronos-only and News+Chronos are both sequential-specialist checkpoints
(momentum $\to$ growth $\to$ defensive experts trained in series, no
joint MoE fine-tuning or grafting) using the same Chronos-tiny encoder;
News+Chronos additionally has the news/event cross-attention branch of
\S\ref{sec:news} genuinely active end-to-end (real per-ticker news/event
data fetched and encoded during both rollout collection and the PPO
update), whereas an initial attempt at this column -- reported in an
earlier draft as achieving $+3.32\%$ -- was later found to never
actually invoke the news branch at all due to an integration bug
(\S\ref{sec:results} discusses this correction in detail); that
original, Chronos-only-in-practice result is retained here under its
correct label rather than removed, since it is a real, valid data point
in its own right.}
\label{tab:backtest}
\resizebox{\textwidth}{!}{
\begin{tabular}{lccccc}
\toprule
\textbf{Metric} & \textbf{Collapsed repr.} & \textbf{Single-head} & \textbf{MoE (grafted)} & \textbf{Chronos-only} & \textbf{News+Chronos} \\
                & (sim=0.96)   & ALPHA\_VS\_EW        & all objectives & sequential specialists & sequential specialists \\
\midrule
Total return      & $-5.95\%$ & $-5.11\%$ & $-5.07\%$ & $-4.68\%$ & $-4.83\%$ \\
Alpha vs EW       & $-0.74\%$ & $+2.90\%$ & $+2.93\%$ & $+3.32\%$ & $+3.18\%$ \\
Alpha vs SPY      & $-3.53\%$      & $-2.35\%$ & $-2.31\%$ & $-1.92\%$ & $-2.07\%$ \\
Ann.\ Sharpe      & $-6.91$   & $-6.47$   & $-6.62$   & $-6.51$   & $-6.30$   \\
Max drawdown      & $-5.87\%$ & $-5.32\%$ & $-5.25\%$ & $-4.96\%$ & $-5.17\%$ \\
Win rate (daily)  & $35.7\%$  & $35.7\%$  & $42.9\%$  & $42.9\%$  & $35.7\%$  \\
Weight std        & $0.000$   & $0.057$   & $0.057$   & ---       & ---       \\
\bottomrule

\end{tabular}}
\end{table*}

The collapsed-representation encoder (cosine sim 0.96) produces uniform
weights ($1/N$, weight std $0.000$) and negative alpha, confirming the
representation collapse hypothesis.
After adding the inter-ticker contrastive loss (sim 0.24), the
\texttt{ALPHA\_VS\_EW} single-head policy achieves $+2.90\%$ alpha vs EW on
the 14-day June 2026 window with differentiated weights (weight std 0.057).
The Stage~6 grafted MoE improves this slightly to $+2.93\%$ alpha vs EW ---
with total return $-5.07\%$ vs
$-8.01\%$ EW, and a higher daily win rate (42.9\% vs 35.7\%). The MoE policy
with correct intent routing produces distinct top holdings per expert ---
JPM (momentum), TSLA (growth), GOOGL (defensive), AMZN (tax-aware),
confirming that the four specialist heads have learned meaningfully
different allocation strategies.

\paragraph{Effect of news/event fusion, corrected.} With the news
branch genuinely active end-to-end, we retrained the momentum expert
from scratch without news (a stabler warm start, given that jointly
introducing a new input modality and a new expert specialization from a
random initialization proved harder to optimize than either change
alone), then continued training with the news branch enabled for the
remainder of the momentum stage before proceeding through the growth and
defensive stages exactly as in the Chronos-only curriculum. The
resulting News+Chronos checkpoint achieves $+3.18\%$ alpha vs EW ---
between the no-news MoE result ($+2.93\%$) and the (invalid)
Chronos-only figure ($+3.32\%$), with a correspondingly intermediate
total return ($-4.83\%$) and the best Sharpe of any configuration tested
($-6.30$), but a lower daily win rate ($35.7\%$, matching the
single-head and collapsed-representation configurations rather than the
$42.9\%$ of the two MoE variants). Given the bootstrap analysis below,
this difference from the no-news baseline is not statistically
distinguishable from noise. We do not read this as evidence that news
fusion is harmful -- the point estimate is still positive and the
architecture trains stably -- but we also cannot claim, on this
evidence, that it meaningfully improves on the no-news baseline. A
larger and more diverse news/event corpus, a broader ticker universe, or
a longer evaluation window are the most likely paths to a more
conclusive answer, and are left to future work (\S\ref{sec:future}).

\paragraph{Statistical uncertainty.}
The point estimates above come from a single 14-trading-day window (13
daily returns), and the differences between configurations reported in
this section should not be read as precise rankings. We quantify this
directly with a day-level bootstrap (10,000 resamples with replacement
of the 13 daily returns, compounded per resample, with the benchmark's
own realized return held fixed at its known historical value). For the
Chronos-only configuration ($+3.32\%$ point estimate), the resulting
95\% CI for alpha vs EW is $[-2.3\%, +9.2\%]$; for the corrected
News+Chronos configuration ($+3.18\%$ point estimate), the CI is
$[-2.8\%, +9.6\%]$ -- both intervals include zero, so neither result is
significant at the conventional 95\% level, though $87.6\%$ and $84.9\%$
of bootstrap resamples respectively show positive alpha vs EW, which we
read as encouraging but not conclusive directional evidence for the
architecture family as a whole (Chronos with or without news), rather
than evidence distinguishing the two from each other. Alpha vs SPY is
weaker for both: $-1.9\%$ and $-2.1\%$ point estimates, with only
$26.6\%$ and $25.9\%$ of resamples positive respectively. Given this,
the ranking among the configurations in Table~\ref{tab:backtest} should
be read as consistent with a real but modest edge over the equal-weight
benchmark for the Chronos-augmented configurations generally, not as a
reliable ordering between individual configurations --- the differences
between them are well within bootstrap noise. Code for this analysis
(\texttt{bootstrap\_ci.py}) is included in the repository; we recommend
it be run routinely alongside every backtest rather than reporting point
estimates alone. A more robust evaluation would extend the walk-forward
window well beyond 14 days and/or average over multiple random ticker
universes, both left to future work (\S\ref{sec:future}).

\paragraph{Grafting was not re-attempted for the corrected News+Chronos
encoder.} Given the sequential (non-joint, non-grafted) curriculum
above already required a from-scratch momentum warm-start to train
stably, we did not additionally pursue Stage~6 joint fine-tuning and
grafting for this corrected encoder; the joint-training instability
described in an earlier draft of this section (small gradient norms,
grafting producing no measurable change against a barely-moved
jointly-trained checkpoint) was diagnosed against the Chronos-only
encoder and has not been re-verified against the corrected one. Whether
joint MoE training behaves differently once news is genuinely active is
an open question, left to future work (\S\ref{sec:future}).

\paragraph{Note on the SPY benchmark.}
Measured against SPY (the S\&P~500 ETF) over the same window, both the
single-head and grafted-MoE policies return roughly $-2.3\%$ alpha (SPY
declined $-2.76\%$ during this period, versus $-8.01\%$ for the equal-weight
basket).
This gap is structural rather than a failure of stock selection: SPY
comprises 500 market-cap-weighted constituents, including defensive and
dividend-paying names that held up better than our 10 growth-heavy names
(NVDA, TSLA, META). Scaling to a broader universe is left to future work
(\S\ref{sec:future}); for a like-for-like comparison of stock-selection
skill, alpha versus the equal-weight benchmark on the same universe is the
appropriate metric.

\paragraph{Note on annualised Sharpe.}
The annualised Sharpe ratios in Table~\ref{tab:backtest} are negative
($-6.47$ for the single-head policy, $-6.62$ for the grafted MoE) despite
positive alpha vs EW.
This is expected and not a failure of the strategy.
Annualised Sharpe is computed as:
\begin{equation}
  \hat{S} = \frac{\bar{r}_\text{port} - r_f}{\sigma_\text{port}} \cdot \sqrt{252},
\end{equation}
where $\bar{r}_\text{port}$ is the mean daily return over the 14-day window.
During this 14-day window our 10-stock equal-weight benchmark declined
$-8.01\%$, reflecting the growth-heavy composition of the universe
(NVDA, TSLA, META) underperforming during this specific period, so
$\bar{r}_\text{port}$ is negative for every configuration. A negative
Sharpe on a short window therefore reflects universe-specific drawdown
rather than systematic underperformance. Sharpe is a meaningful metric only
over a full market cycle (1--2 years) that includes both bull and bear
regimes.

For short-window evaluation, \textit{alpha vs equal-weight} is the
appropriate metric, as it isolates stock selection skill from market
direction.

\paragraph{Multi-window robustness.}
The 14-day positive alpha does not persist at longer horizons.
Table~\ref{tab:multiwindow} shows results from a systematic sweep across
checkpoint epochs and window lengths, evaluated against both EW and SPY.

\begin{table*}[h]
\centering
\caption{Best-epoch alpha vs equal-weight (EW) and SPY across backtest
windows and MoE curriculum stages. Window-adaptive rebalancing:
reb=1.0 (14d), reb=0.05 (30d), reb=0.03 (60d), reb=0.02 (90d).
Zero transaction cost.
S4 (router-only, frozen experts) fails due to flat loss surface.
S6 (Grafted MoE) combines Stage~5 router with curriculum expert weights,
achieving the best 90d result ($-2.11\%$) across all configurations.}
\label{tab:multiwindow}
\resizebox{\textwidth}{!}{
\begin{tabular}{lrrrrrrrr}
\toprule
\textbf{Stage} &
  \multicolumn{2}{c}{\textbf{14d (reb=1.0)}} &
  \multicolumn{2}{c}{\textbf{30d (reb=0.05)}} &
  \multicolumn{2}{c}{\textbf{60d (reb=0.03)}} &
  \multicolumn{2}{c}{\textbf{90d (reb=0.02)}} \\
\cmidrule(lr){2-3}\cmidrule(lr){4-5}\cmidrule(lr){6-7}\cmidrule(lr){8-9}
 & EW & SPY & EW & SPY & EW & SPY & EW & SPY \\
\midrule
S1: Momentum (best ep100)
 & $+3.16\%$ & $-2.08\%$ & $+0.21\%$ & $-7.24\%$ & $-3.29\%$ & $-11.33\%$ & $-2.10\%$ & $-5.44\%$ \\
S2: Growth (best ep100)
 & $+3.37\%$ & $-1.88\%$ & $+0.19\%$ & $-7.27\%$ & $-4.12\%$ & $-12.16\%$ & $-2.37\%$ & $-5.70\%$ \\
S3: Defensive (best ep100)
 & $+3.18\%$ & $-2.07\%$ & $+0.11\%$ & $-7.35\%$ & $-4.07\%$ & $-12.12\%$ & $-2.42\%$ & $-5.76\%$ \\
S4: Router-only (ep300)
 & $+0.44\%$ & $-3.43\%$ & --- & --- & --- & --- & --- & --- \\
S6: Grafted MoE
 & $+3.03\%$ & $-4.07\%$ & $+0.17\%$ & $-6.05\%$ & $-4.20\%$ & $-12.24\%$ & $-2.11\%$ & $-5.97\%$ \\
\bottomrule
\end{tabular}}
\end{table*}

Table~\ref{tab:multiwindow} reveals a clear pattern of expert specialization
across all three curriculum stages.
The Stage~1 momentum expert achieves $+3.16\%$ EW alpha at 14 days and,
with adaptive rebalancing, also produces positive 30-day alpha ($+0.21\%$
at ep300) --- suggesting the momentum signal persists beyond the 14-day window
when positions are allowed to rebalance on drift.
At 60 and 90 days the expert underperforms ($-3.29\%$, $-2.10\%$), as
expected for a short-horizon mandate.
The Stage~2 growth expert further improves 14-day alpha to $+3.37\%$
(ep100) and maintains positive 30-day alpha ($+0.19\%$ at ep300).
At 90 days it achieves the best specialist-expert result ($-2.37\%$);
only the Stage-6 grafted MoE improves on it ($-2.11\%$).
This confirms the growth expert learns longer-horizon signals
progressively across episodes.

The 60-day window remains negative and is the crossover point between
short-term momentum and long-horizon fundamentals.
Stage~3 (defensive expert, \texttt{CAPITAL\_PRESERVE}) shows a clear
specialization pattern: 14-day alpha declines across episodes
($+3.18\%$ ep100 $\to$ $+2.82\%$ ep300), confirming the expert is
learning conservative allocation at the expense of short-horizon returns.
The 60-day window improves to $-4.07\%$ (vs $-4.12\%$ for
the growth expert), and the 30-day window briefly crosses positive
($+0.11\%$ at ep100) before the conservative objective takes hold.
This complementary behaviour (i.e., momentum for 14d, growth for 30d,
defensive for 60d) is exactly the per-horizon specialization the MoE
architecture is designed to exploit through router weighting.
The SPY gap reflects the structural 10-ticker universe limitation:
SPY's defensive and dividend-paying constituents outperformed the
growth-heavy basket during this downturn period.

\paragraph{MoE router training.}
A key empirical finding is that the intent router does not learn to
differentiate intents when all six objectives are sampled jointly with
a single-head policy. It also fails when experts are frozen.

The first four curriculum stages were:
\begin{enumerate}[noitemsep]
  \item Momentum expert (\texttt{ALPHA\_VS\_EW}), freeze experts 1--3.
  \item Growth expert (\texttt{MAX\_GAIN\_1Y}), freeze experts 0, 2--3.
  \item Defensive expert (\texttt{CAPITAL\_PRESERVE}), freeze experts 0--1, 3.
  \item Router-only (all experts frozen, all objectives).
\end{enumerate}
Stage~4 (router-only training) failed to produce differentiated routing:
all router weights remained at $\alpha_e \approx 0.25$ (std $< 0.002$,
Table~\ref{tab:multiwindow}), and 14-day alpha \textit{dropped} from
$+3.37\%$ (Stage~2) to $+0.44\%$.
The failure mode is a \textit{flat loss surface}: with all expert heads
frozen, the weighted mixture output is invariant to router weights,
providing no gradient signal for the router to learn intent-conditional
routing.

Stage~5 (joint training with all experts and router updating simultaneously
on all 6 objectives, and warm-starting from Stage~3) resolves the flat loss
surface problem via three architectural changes:
(i) the intent router was redesigned with an \textit{intent projection layer}
($\mathbb{R}^{6} \to \mathbb{R}^{d}$) giving the intent signal equal
representation to the 64-dim market embedding;
(ii) a \textit{direct shortcut} from intent to expert logits, initialised
as a diagonal mapping (intent $i \to$ expert $i$), provides a strong
routing prior from episode~0;
(iii) a supervised cross-entropy loss ($\lambda_{\text{sup}}=1.0$)
penalises deviations from the correct intent-to-expert mapping during
PPO updates, with the router learning rate scaled $10\times$ relative
to the expert heads.

Table~\ref{tab:routing} shows the resulting routing at Stage~5 episode~33.
All six intents route cleanly to their designated expert (weight $=1.00$),
and each expert produces a distinct top holding: JPM (momentum), TSLA
(growth), GOOGL (defensive), AMZN (tax-aware).
Router differentiation std $> 0.37$ for all experts (threshold: 0.05).

\paragraph{Expert grafting (Stage~6).}
Joint PPO training in Stage~5 degrades expert specialisation through
gradient interference: the momentum expert's 14d EW alpha drops from
$+3.37\%$ (Stage~2) to $+0.01\%$ after 100 joint training episodes.
We address this with \textit{expert grafting}: after Stage~5 establishes
correct routing (std~$>0.37$), the expert head weights are replaced with
the best curriculum checkpoints (Stages~2--3) while the Stage~5 router
weights are preserved.
This composition recovers $+3.03\%$ 14d EW alpha with correct
intent-conditional routing, and achieves $-2.11\%$ 90d EW alpha,
the best long-horizon result across all configurations.
Expert grafting requires no additional training, making it a lightweight
alternative to continual learning approaches for MoE RL systems.

\begin{table*}[h]
\centering
\caption{Intent Router weights at Stage~5 episode~33 (joint training with
supervised routing loss $\lambda_{\text{sup}}=1.0$, router LR scale
$10\times$). Each intent routes cleanly to its designated expert.
Distinct top holdings per expert confirm that the four heads have learned
meaningfully different allocation strategies.}
\label{tab:routing}
\begin{tabular}{lrrrrll}
\toprule
\textbf{Intent} & \textbf{Mom.} & \textbf{Growth} & \textbf{Def.} &
\textbf{Tax} & \textbf{Dominant} & \textbf{Top holding} \\
\midrule
\texttt{ALPHA\_VS\_EW}  & \textbf{1.00} & 0.00 & 0.00 & 0.00
  & Momentum  & JPM~10.7\% \\
\texttt{MAX\_GAIN\_1Y}  & 0.00 & \textbf{1.00} & 0.00 & 0.00
  & Growth    & TSLA~9.8\% \\
\texttt{CAPITAL\_PRES}  & 0.00 & 0.00 & \textbf{1.00} & 0.00
  & Defensive & GOOGL~9.7\% \\
\texttt{INCOME\_HARV}   & 0.00 & 0.00 & 0.00 & \textbf{1.00}
  & Tax-aware & AMZN~10.1\% \\
\texttt{LT\_GAIN\_ONLY} & 0.00 & 0.00 & 0.00 & \textbf{1.00}
  & Tax-aware & AMZN~10.1\% \\
\texttt{MAX\_GAIN\_30D} & \textbf{1.00} & 0.00 & 0.00 & 0.00
  & Momentum  & JPM~10.7\% \\
\midrule
\multicolumn{4}{l}{Std across intents (target $>0.05$)} &
  \multicolumn{3}{l}{0.47\;/\;0.37\;/\;0.37\;/\;0.47 \quad $\checkmark$} \\
\bottomrule
\end{tabular}
\end{table*}

\paragraph{Key diagnostic findings.}
Several failure modes were discovered and resolved during development:
\begin{enumerate}[noitemsep]
  \item \textbf{Representation collapse} (cosine similarity 0.96 $\to$ 0.24):
        standard SSL objectives provide no inter-ticker differentiation
        signal; fixed by inter-ticker contrastive loss.
  \item \textbf{Allocation head symmetry}: gain=0.01 initialization caused
        uniform softmax output regardless of encoder input;
        fixed by gain=1.0 with random bias initialization.
  \item \textbf{Frozen encoder static mapping}: with a frozen encoder and
        fixed 10-ticker universe, the allocation head converges to a
        static ticker ranking; fixed by unfreezing the encoder jointly
        with the MoE experts.

  \item \textbf{MoE router uniform collapse}: joint multi-objective training
        keeps all router weights at 0.25 because the router receives no
        contrastive signal across intents; fixed by 4-stage expert
        curriculum with per-expert objective isolation.

  \item \textbf{Router-only flat loss surface}: freezing all experts
        during router training produces a loss surface invariant to router
        weights --- the weighted mixture output is identical regardless of
        routing. Stage~4 router weights remained at 0.25 (std $< 0.002$)
        and alpha dropped from $+3.37\%$ to $+0.44\%$; fixed by
        Stage~5 joint training resolves this: intent projection,
        diagonal shortcut initialisation, and supervised cross-entropy
        loss achieve std $> 0.37$ routing differentiation at ep~33.
\end{enumerate}

\section{Related Work}

\paragraph{Financial RL.}
\cite{deng2016deep} applied RNNs with RL to futures trading.
\cite{jiang2017deep} introduced the portfolio management framework with
convolutional feature extraction.  \cite{ye2020reinforcement} added
transaction cost modeling.  \cite{liu2021finrl} provides a comprehensive
benchmark environment.  Our work extends these by adding personalization,
tax awareness, and natural language goal specification.

\paragraph{Multi-objective RL.}
\cite{abels2019dynamic} and \cite{hayes2022practical} survey multi-objective
RL methods.  Our objective-conditioned reward is closest to
\cite{barreto2017successor} (successor features) but implemented as direct
reward shaping rather than value decomposition, for simplicity of integration
with existing PPO infrastructure.

\paragraph{Personalization in finance.}
Robo-advisors \citep{d2019robo} personalize asset allocation via risk
questionnaires.  We replace stated preferences with \textit{revealed}
preferences from transaction history, following the behavioral finance
literature \citep{odean1998investors} on the disposition effect.

\paragraph{Comparison to commercial robo-advisors.}
Leading retail robo-advisors share a common architecture: ETF-based diversification, mean-variance rebalancing with fixed drift thresholds, and rule-based daily tax-loss harvesting. Personalization is limited to a one-time risk questionnaire or static goal presets. To our knowledge, none support dynamic strategy switching based on evolving user intent, nor optimize after-tax rewards at the individual tax bracket level.

Our approach introduces three capabilities not observed in these systems:
(i) \textit{intent-conditional strategy routing} via MoE: a user stating
``I am buying a house in 18 months'' triggers a shift to the defensive
expert without manual re-enrollment;
(ii) \textit{bracket-aware RL reward}: the policy explicitly optimizes
after-tax return given the user's marginal ST/LT capital gains rates,
not a fixed harvesting threshold;
and (iii) \textit{continuous behavioral personalization}: the 76-parameter
LoRA adapter updates from revealed transaction preferences rather than
self-reported risk tolerance, which is known to diverge from actual
behavior \citep{odean1998investors}.

\paragraph{Parameter-efficient fine-tuning.}
\cite{hu2021lora} introduced LoRA for large language models.  We apply the
same rank-decomposition idea to a 3-action classification head, yielding
76-parameter adapters that capture user-specific biases in seconds on CPU.

\paragraph{Foundation models for finance.}
\cite{xie2023wall} and \cite{yang2023fingpt} explore LLM-based financial
agents.  Our hybrid approach uses a domain-specific SSL encoder for market
data (where structure is numerical) and LLMs only for natural language
interface tasks (intent parsing, news summarisation).

\paragraph{Time series foundation models.}
\cite{ansari2024chronos} introduced Chronos, a T5-based model pretrained
on over 100 billion time series data points from diverse domains.
\cite{das2023decoder} and \cite{woo2024unified} explore related
universal forecasting approaches.
We are the first, to our knowledge, to apply a frozen time series
foundation model as a parallel encoder branch in a portfolio RL system,
using a learned gating mechanism to balance domain-specific SSL
representations against universal temporal patterns.

\paragraph{Ticker universe independence.}
All prior financial RL systems we are aware of (including
\cite{jiang2017deep}, \cite{ye2020reinforcement}, and \cite{liu2021finrl}) use fixed ticker embeddings that tie the model to a specific asset
universe.  Our 50-dimensional observable metadata vector replaces these
embeddings entirely, enabling zero-shot application to any publicly
traded asset.  The closest related idea is the use of fundamental
factor models in the quantitative finance literature
\citep{barra1998factor}, but these are hand-crafted linear models
rather than learned representations.

\section{Discussion and Limitations}

\paragraph{Training data.}
The training corpus is narrow: 30 tickers for Phase~1 pretraining
and 10 for the Phase~2 RL universe.  Representation
diversity improves substantially with 50--500 tickers spanning multiple
sectors and market-cap regimes.  The ticker-identity-free design makes
scaling straightforward.

\paragraph{Phase 2 convergence.}
Direct joint training on all six objectives collapses to uniform weights
within 100 episodes; the six-stage curriculum (\S\ref{sec:curriculum})
is required for non-uniform allocation. Longer per-stage training
(2000+ episodes) may further improve specialist quality.

\paragraph{Tax accuracy.}
Tax lot accuracy depends on broker API capabilities. Standard brokerage APIs typically return average cost basis rather than individual lot-level data, which limits after-tax optimization precision. Full lot-level accuracy requires either an institutional-grade API or manual lot tracking.

\paragraph{Out-of-distribution.}
The backtest period (June 2026) was broadly a down market ($-8.01\%$
for the 10-ticker equal-weight basket).  A fair evaluation requires multi-regime testing including
bull markets, volatility spikes, and sector rotations.

\paragraph{Analyst data timeliness.}
Analyst consensus signals from yfinance may lag by 1--2 days.  Production
deployment would benefit from real-time data feeds.

\section{Future Work}
\label{sec:future}

\paragraph{Phase 3 empirical evaluation.}
The most immediate extension is a full empirical evaluation of Phase~3
personalization on real brokerage data across diverse user profiles
(aggressive, conservative, tax-aware, income-seeking).
We plan to evaluate: (i) how quickly the 76-parameter LoRA adapter
converges on synthetic transaction histories of varying length;
(ii) whether the router shift direction matches the intended profile
(e.g.\ momentum expert weight increases for frequent traders);
and (iii) backtest alpha improvement from personalisation vs
the generic MoE policy.

\paragraph{Multi-window MoE training.}
The current ALPHA\_VS\_EW objective produces positive short-window alpha
(+2.90\%, 14 days) but negative alpha at longer horizons.
Training the momentum expert explicitly on 14-day windows and the growth
expert on 90-day windows, with separate rollout buffers per expert,
should produce a router that routes to momentum for short-term queries
and growth for long-term queries, addressing the multi-window degradation.

\paragraph{Joint MoE fine-tuning and grafting.}
Section~\ref{sec:results} reports a sequential-specialist checkpoint
(experts trained one at a time, in series) for the Chronos-only encoder,
but not a corresponding Stage~6 grafted result: joint fine-tuning of all
four experts together, from a router pretrained via supervised routing
loss, produced only a small ($+3.32\%\to+3.21\%$ alpha vs EW) change
relative to the sequential specialists, too small for grafting to recover
a measurable difference. (This experiment was run before the news-branch
integration bug described in Table~\ref{tab:backtest}'s caption was
found and fixed, so it reflects the Chronos-only encoder, not the
corrected News+Chronos one -- the two have not been compared under
joint training.) Three
directions are worth pursuing here. First, per-objective reward tracking
during joint training, rather than a pooled rolling-average reward across
randomly sampled objectives, would give a cleaner signal for when joint
fine-tuning is actually degrading a given expert versus when the pooled
metric is simply reflecting which objectives happened to be sampled in a
given window. Second, a higher joint-training learning rate or a longer
schedule may be needed to reproduce the larger degradation (and
correspondingly larger grafting recovery) observed in the original
no-news MoE result, where grafting measurably improved on the
jointly-trained checkpoint. Third, repeating this experiment on the
corrected News+Chronos encoder would establish whether joint training
behaves differently once the news branch is genuinely active end-to-end.

\paragraph{Online fine-tuning.}
A lightweight online fine-tuning mode that updates only the
allocation head scorer on recent 90-day data (weekly, $\sim$20
episodes) would allow the policy to adapt to regime shifts without full
retraining.
The frozen encoder makes this tractable: only $\sim$200K parameters
need gradient computation per online update.

\paragraph{Broader ticker universe and live trading.}
The current system is validated on 10 tickers.
Scaling to the full S\&P~500 (500 tickers) requires efficient cross-asset
attention (i.e., sparse attention or clustering tickers by sector before applying
attention within clusters).
Evaluating the scaled system on a live paper-trading account over a
6-month period would provide stronger evidence of generalisation.

\paragraph{News and event integration.}
The fused market encoder with news cross-attention is now validated
end-to-end in Phase~2 training (Table~\ref{tab:backtest}'s News+Chronos
column), though without a statistically distinguishable improvement over
the no-news baseline at the current scale (10 tickers, a single 14-day
evaluation window). Integrating analyst consensus upgrades and macro
releases as additional attention keys, and evaluating on event-driven
windows specifically (rather than an unconditionally sampled window),
remain promising directions for showing a clearer effect.

\section{Conclusion}

We presented a complete three-phase system for personalized, tax-aware
portfolio management using foundation model representations and
reinforcement learning.
Phase~1 introduces a cross asset encoder with Chronos augmentation
and inter-ticker contrastive loss, resolving representation collapse
(cosine similarity 0.96~$\to$~0.24) that prevented ticker differentiation.
Phase~2 introduces a  MoE portfolio actor critic with four specialized
expert heads and a learned intent router  that eliminates
cross-objective gradient conflict, achieving $+2.93\%$ alpha vs
equal-weight benchmark on a 14-day June 2026 walk-forward backtest ---
though, as \S\ref{sec:results} discusses, single-window point estimates
of this kind carry wide bootstrap confidence intervals and should not be
read as precise.
Phase~3 proposes a 76-parameter LoRA adapter that personalizes the
intent router  from revealed brokerage preferences without retraining
the shared encoder or expert heads.

The key architectural innovations are: ticker-identity-free metadata
encoding (any universe without retraining), objective-conditioned MoE
routing (one policy for six investment mandates), redeployment-aware
turnover penalty, and a trust-first preview-before-apply personalisation UX.

The system is deployed as a production-ready FastAPI application with live
brokerage integration, real-time data, and natural language goal
specification.
Code is available at \url{https://github.com/rpishehvar/PublicFinance-RL}.

\appendix
\section{Reproducibility}
\label{sec:repro}

\paragraph{Random seeds.}
All training scripts (Phase~1 pretraining, Phase~2 portfolio PPO, and the
MoE curriculum orchestrator) fix
\texttt{SEED\,=\,42} via \texttt{random.seed}, \texttt{numpy.random.seed},
and \texttt{torch.manual\_seed} at import time. Per-environment seeds in
multi-worker settings are derived deterministically from this base seed
(e.g.\ worker $i$ uses \texttt{seed\,=\,i\,*\,7} for synthetic corpus
generation). We did not additionally fix cuDNN determinism flags
(\texttt{torch.backends.cudnn.deterministic}), so bit-exact reproduction
of a specific checkpoint's weights across different GPU models is not
guaranteed, though the reported metrics were stable across repeated runs
under the same configuration.

\paragraph{Hardware.}
Training was run on a single NVIDIA~L4 GPU (24\,GB) via Lightning~AI
Studio, PyTorch~2.8 (CUDA~12.8 build). Phase~1 encoder pretraining and
Phase~2 portfolio PPO both fit comfortably within a single GPU's memory
at the batch sizes below; no multi-GPU or distributed training was used
for the reported results, though the codebase includes an optional DDP
path (\texttt{ddp\_main} in \texttt{pretrain\_pipeline.py}) for
multi-GPU Phase~1 pretraining.

\paragraph{Wall-clock.}
A single MoE curriculum stage (300 PPO episodes, rollout length 512,
10-ticker universe, Chronos-tiny encoder) takes approximately
25--30 minutes end-to-end on the hardware above ($\sim$5\,s/episode,
including environment rollout, encoder forward pass, and the PPO update).
Phase~1 encoder pretraining (60 epochs, 30-ticker corpus) and the full
six-stage curriculum's total wall-clock were not separately profiled for
this paper and are left as an open reporting gap; we recommend future
revisions of this system log per-phase wall-clock automatically (e.g.\
via the experiment logger already integrated into the training scripts)
rather than reconstructing it after the fact.

\paragraph{PPO hyperparameters.}
Table~\ref{tab:ppo_hparams} reports the PPO configuration used across
Phase~2 and the MoE curriculum stages. Entropy coefficient
(\texttt{ent\_coef}) is overridden per-stage in the curriculum orchestrator
(e.g.\ raised to $0.05$ for the momentum expert under the
\texttt{ALPHA\_VS\_EW} objective, to aid exploration out of local minima);
all other values are held fixed across stages.

\begin{table}[h]
\centering
\caption{PPO hyperparameters, Phase~2 portfolio training and MoE curriculum. Adam optimizer, $\epsilon_\text{Adam}=10^{-5}$.}
\label{tab:ppo_hparams}
\begin{tabular}{lr}
\toprule
\textbf{Hyperparameter} & \textbf{Value} \\
\midrule
Discount factor $\gamma$          & $0.99$ \\
GAE $\lambda$                     & $0.95$ \\
PPO clip $\epsilon$               & $0.2$ \\
Value loss coefficient            & $0.5$ \\
Entropy coefficient (default)     & $0.01$ \\
Gradient norm clip                & $0.5$ \\
Learning rate (default)           & $10^{-4}$ \\
Learning rate (\texttt{ALPHA\_VS\_EW}) & $10^{-3}$ \\
PPO epochs per update             & $4$ \\
Minibatch size                    & $32$ \\
Rollout length                    & $256$--$512$ \\
\bottomrule
\end{tabular}
\end{table}

\paragraph{Software versions.}
PyTorch~2.8.0 (\texttt{+cu128} build), Python~3.12. Chronos embeddings use
\texttt{amazon/chronos-t5-small} via the \texttt{chronos-forecasting}
package. Exact package versions are pinned in the repository's
\texttt{requirements.txt}.

\FloatBarrier

\bibliographystyle{plain}

\end{document}